\documentclass[10pt,twocolumn,letterpaper]{article}

\usepackage{cvpr}
\usepackage{times}
\usepackage{epsfig}
\usepackage{graphicx}
\usepackage{amsmath}
\usepackage{amssymb}
\usepackage{graphics}
\usepackage{booktabs}
\usepackage{multirow}
\usepackage{enumitem}
\usepackage{bm}
\usepackage[dvipsnames,table]{xcolor}
\usepackage{adjustbox}
\usepackage[innercaption]{sidecap}
\usepackage{upquote}
\usepackage{textcomp}
\usepackage{pifont}
\usepackage{natbib}
\usepackage{etoolbox}
\usepackage{caption}
\usepackage{subcaption}
\usepackage[utf8]{inputenc}
\usepackage{xspace}
\usepackage{enumitem}
\usepackage{wasysym}
\usepackage{lipsum}
\usepackage{pict2e}
\usepackage{relsize}
\usepackage{soul}
\usepackage{marvosym}

\sidecaptionvpos{figure}{m}
\graphicspath{{./}}
\setcitestyle{numbers,square,citesep={,}}
\DeclareGraphicsExtensions{.jpeg,.png,.jpg,.pdf}
\newcommand{\Ree}{\mathbb{R}}

\newcommand{\timesnarrow}{{\mkern-2mu\times\mkern-2mu}}
\newcommand{\CheckBoxCustom}{\makebox[0pt][l]{$\square$}\raisebox{.15ex}{\hspace{0.1em}$\boldsymbol{\checkmark}$}}
\newcommand{\partitle}[1]{\noindent\textbf{#1}}
\newcommand{\ptspace}{\vspace*{5pt}}

\DeclareFontFamily{U}{mathb}{\hyphenchar\font45}
\DeclareFontShape{U}{mathb}{m}{n}{
<-6> mathb5
<6-7> mathb6
<7-8> mathb7
<8-9> mathb8
<9-10> mathb9
<10-12> mathb10
<12-> mathb12}{}
\DeclareSymbolFont{mathb}{U}{mathb}{m}{n}
\DeclareMathSymbol{\llcurly}{\mathrel}{mathb}{"CE}
\DeclareMathSymbol{\ggcurly}{\mathrel}{mathb}{"CF}

\definecolor{actioncolor01}{HTML}{953735}
\definecolor{actioncolor02}{HTML}{997800}
\definecolor{actioncolor03}{HTML}{4F6228}
\definecolor{actioncolor04}{HTML}{9B540D}
\definecolor{actioncolor05}{HTML}{376092}
\definecolor{actioncolor06}{HTML}{604A7B}

\definecolor{actioncolor11}{HTML}{F2DCDB}
\definecolor{actioncolor12}{HTML}{FFEDB4}
\definecolor{actioncolor13}{HTML}{D7E4BD}
\definecolor{actioncolor14}{HTML}{FFDFB5}
\definecolor{actioncolor15}{HTML}{DCE6F2}
\definecolor{actioncolor16}{HTML}{E6E0EC}

\definecolor{lightgray}{gray}{0.6}
\cvprfinalcopy

\begin{document}

\title{TimeGate: Conditional Gating of Segments in Long-range Activities}

\author{Noureldien Hussein$^{\text{1,2}}$\footnotemark
\\
$^{\text{1}}$University of Amsterdam \\
\tt\small{nhussein@uva.nl}
\and
Mihir Jain$^{\text{2}\dag}$ \hspace*{10pt} \qquad Babak Ehteshami Bejnordi$^{\text{2}\dag}$ \\
$^{\text{2}}$Qualcomm AI Research, Qualcomm Technologies Netherlands B.V. \\
\tt\small{\{mijain, behtesha\}@qti.qualcomm.com}  \qquad}

\maketitle

\renewcommand*{\thefootnote}{\fnsymbol{footnote}}
\setcounter{footnote}{1}
\footnotetext{Research conducted during an internship at Qualcomm AI Research.}
\setcounter{footnote}{2}
\footnotetext{Qualcomm AI Research is an initiative of Qualcomm Technologies, Inc.}
\renewcommand*{\thefootnote}{\arabic{footnote}}
\setcounter{footnote}{0}

\begin{abstract}
When recognizing a long-range activity, exploring the entire video is exhaustive and computationally expensive, as it can span up to a few minutes.
Thus, it is of great importance to sample only the salient parts of the video.
We propose TimeGate, along with a novel conditional gating module, for sampling the most representative segments from the long-range activity.
TimeGate has two novelties that address the shortcomings of previous sampling methods, as SCSampler.
First, it enables a differentiable sampling of segments.
Thus, TimeGate can be fitted with modern CNNs and trained end-to-end as a single and unified model.
Second, the sampling is conditioned on both the segments and their context.
Consequently, TimeGate is better suited for long-range activities, where the importance of a segment heavily depends on the video context.
TimeGate reduces the computation of existing CNNs on three benchmarks for long-range activities: Charades, Breakfast and MultiThumos. In particular, TimeGate reduces the computation of I3D by 50\% while maintaining the classification accuracy.

\end{abstract}

\section{Introduction}
\label{sec:introduction}
A human can skim through a minute-long video in a few seconds, and still grasp its underlying story~\cite{szelag2004individual}.
This extreme efficiency in temporal information processing raises a question. Can a neural model achieve such efficiency in recognizing minutes-long activities in videos?

Related works propose different CNN models with efficiency in mind~\cite{howard2017mobilenets,zhang2018shufflenet,zoph2018learning}.
However, such models~\cite{kopuklu2019resource} address only short-range actions, as in Kinetics~\cite{kay2017kinetics}, UCF-101~\cite{soomro2012ucf101}, or HMDB~\cite{kuehne2011hmdb}.
On average, these actions take ten seconds or less, where recognizing only a few frames would suffice~\cite{schindler2008action}.
However, this paper focuses on long-range activities, as in Charades~\cite{sigurdsson2016hollywood}, Breakfast~\cite{kuehne2014language} or MultiThumos~\cite{yeung2018every}.
These activities can take up to a few minutes to unfold.
Current methods fully process the entire video of long-range activity to successfully recognize it~\cite{wang2018non,carreira2017quo}.
As a result, the major computational bottleneck of such methods is the sheer number of video frames to be processed.

\begin{figure}[!t]
\begin{center}
\includegraphics[trim=2mm 0mm 1mm 0mm,width=1.0\linewidth]{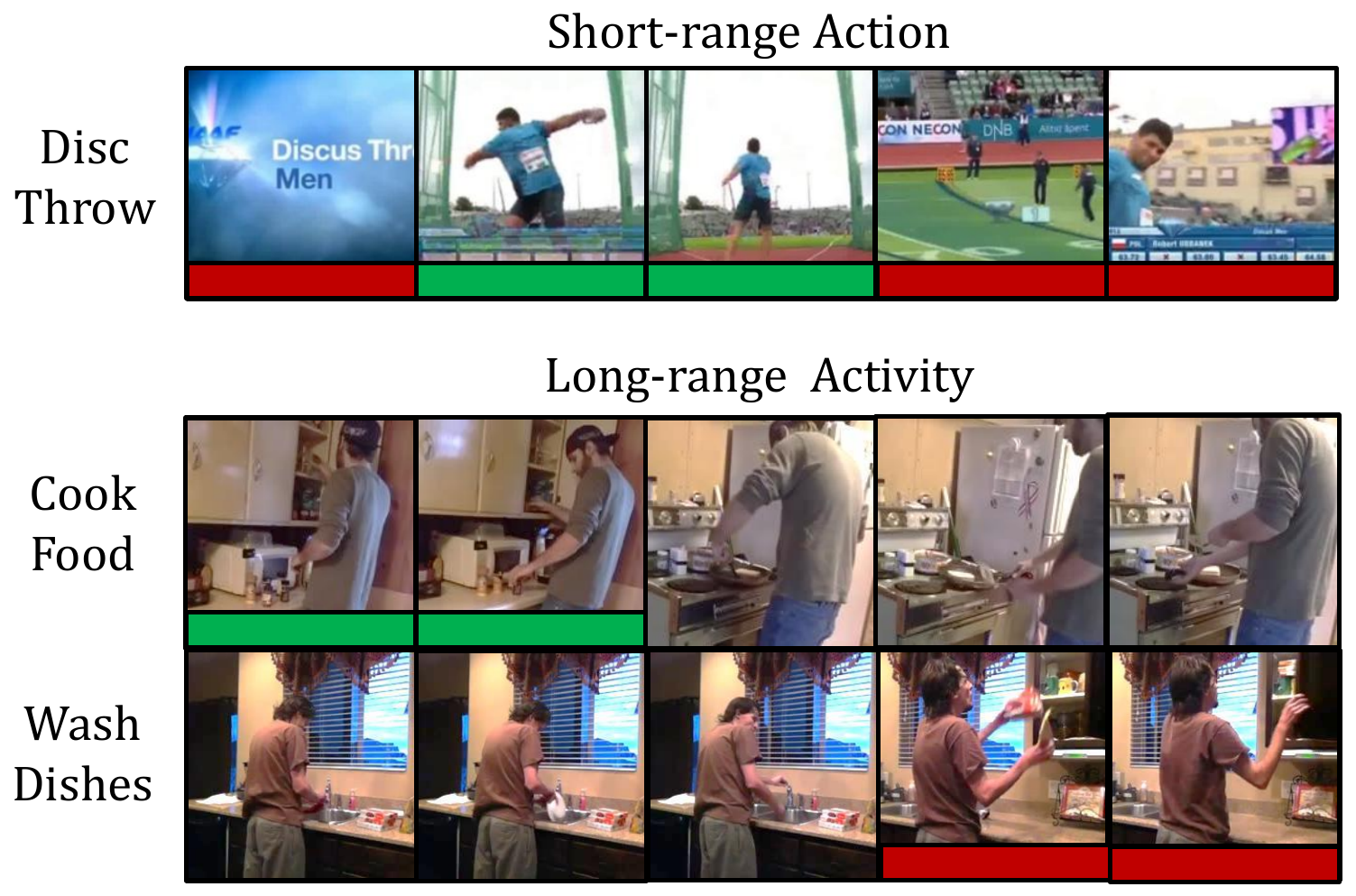}
\end{center}
\vspace*{-5mm}
\caption{Top, short-range action ``disc throw" in an untrimmed video.
Based on each segment, you can tell whether is it relevant (green) to the action or not (red).
But in long-range activities, middle and bottom, the importance of each segment is conditioned on the video context.
The segment ``get food from cupboard" is relevant to ``cook food" but not to ``washing dishes".}
\label{fig:1-1}
\vspace*{-5mm}
\end{figure}

Another solution is frame sampling~\cite{yeung2016end}.
The recently proposed SCSampler~\cite{korbar2019scsampler} achieves efficiency by sampling the most salient segments from an untrimmed video of short-range action.
The sampling is conditioned on only the segment level, which is plausible for short-range actions in trimmed videos, such as Kinetics~\cite{kay2017kinetics} or untrimmed videos, such as Sports-1M~\cite{karpathy2014large}.
The reason is that, on the segment level, one can easily tell if the segment is relevant to the action or it is just background, see figure 1.
So, segment-level classification probabilities would suffice for sampling~\cite{korbar2019scsampler}.
In contrast, long-range activities are known for being diverse and complex~\cite{hussein2019timeception,ye2015eventnet}.
Thus, the importance of one segment to a certain activity is not self-described but rather depends on the context, \textit{i.e.} the long-range activity itself.
That is to say, while a segment is relevant to one activity, it is not relevant to another.
So, sampling conditioned only on the segment level is not the most optimal choice for long-range activities.

To address the limitations of the previous methods, we propose TimeGate, a two-stage neural network for the efficient recognition of long-range activities without compromising the performance.
Different from previous sampling methods, such as SCSampler, TimeGate solves two problems.
\textit{i.} Conditional selection: when selecting segments from the long-range activity, TimeGate is conditioned on both the segment- and context-level features.
Context-conditioning better suited for long-range activities than only the segment-conditioning of SCSampler.
\textit{ii.} Differentiable gating: the selection mechanism of TimeGate is differentiable, so it is trained end-to-end with modern 2D and 3D CNNs~\cite{carreira2017quo,he2016deep}, resulting in a better performance.

Our novelties are:
\textit{i.} Gating module for the conditional sampling of segments in videos.
Our gating is more suited to long-range activities than other methods, such as SCSampler.
The reason is that the sampling is conditioned on the segment- and context-level features.
\textit{ii.} The proposed gating module is differentiable, which enables end-to-end training with existing CNNs.
\textit{iii.} The proposed model, TimeGate, reduces the computational cost of existing CNNs in recognizing long-range activities.
In end-to-end training, the cost is reduced even further.
Finally, we conduct experiments and report the results on three datasets for long-range activity recognition: Charades~\cite{sigurdsson2016hollywood}, Breakfast~\cite{kuehne2014language} and MultiThumos~\cite{yeung2018every}.

\section{Related Work}
\label{sec:related_work}
\partitle{Long-range Activities.}
Short-range actions, such as Kinetics~\cite{kay2017kinetics} and UCF-101~\cite{soomro2012ucf101}, have an average length of 10 seconds or less.
Practically, they can be classified with CNNs using as little as ten frames per video~\cite{wang2016temporal}, and in some cases, even one frame would suffice~\cite{schindler2008action}.
Therefore, building efficient CNNs is a plausible choice to reduce the computational cost of recognizing short-range actions.
However, in long-range activities, such as Charades~\cite{sigurdsson2016hollywood} and Breakfast~\cite{kuehne2014language}, the activity can take up to five minutes to unfold.
Thus, requiring as many as a thousand frames~\cite{hussein2019timeception,hussein2019videograph,hussein2020pic,hussein2017unified} to be correctly classified.
As such, analyzing all the frames using efficient CNNs is still computationally expensive.

Nevertheless, having a mechanism to select the most relevant frames can boost efficiency~\cite{bhardwaj2019efficient}.
Therefore, this paper focuses on reducing the number of video frames needed for activity recognition.
Though, our work is orthogonal to prior work of efficient CNNs for action recognition.

\ptspace
\partitle{Efficient Architectures.}
 CNNs are the go-to solution when it comes to video classification.
Thus, one prospective of reducing the computation of video recognition is to build efficient CNNs.
Methods for pruning less important weights~\cite{hassibi1993optimal,han2015learning} or filters~\cite{li2016pruning} were previously proposed. 
Careful design choices result in very efficient 2D CNNs such as MobileNet~\cite{howard2019searching} and ShuffleNet~\cite{zhang2018shufflenet}.
These 2D CNNs are extended to their 3D counterparts, such as ShuffleNet3D and MobileNet3D~\cite{kopuklu2019resource}, to learn spatio-temporal concepts for video classification.
Neural architecture search~\cite{zoph2016neural} is used to find the lightweight NasNet-Mobile~\cite{zoph2018learning}.

While efficient architectures are successful in the case of short-range actions, they are not the most viable solution for long-range activities.
The reason is that these activities span up to a few minutes.
Naively processing the video in its entirety undermines the computation saved by these efficient CNNs.
In other words, in the case of long-range activities, the computational bottleneck is the sheer number of video segments needed to be processed.

\ptspace
\partitle{Conditional Computing.}
Another solution to reduce the computation is to dynamically route the computational graph of a neural network.
The assumption is that not all input signals require the same amount of computation -- some are complicated while others are seemingly easy.
Thanks to categorical reparameterization~\cite{jang2016categorical}, it becomes possible to discretize a continuous distribution, and effectively learn binary gating.
In~\cite{veit2018convolutional}, a dynamical graph is built by gating the layers of a typical CNN.
While in~\cite{chen2019you,bejnordi2019batch}, the gating is achieved on the level of convolutional channels.
In the same vein, GaterNet~\cite{chen2019you} proposes a separate gating network to learn binary gates for the backbone network.

Rather than gating the network layers, this paper focuses on gating the video frames themselves, to realize the efficiency in recognizing long-range activities.
In all cases, our paper benefits from prior work of differentiable gating~\cite{jang2016categorical}.

\ptspace
\partitle{Sampling of Video Segments.}
Several works discuss frame sampling for short-range videos.
In~\cite{bhardwaj2019efficient}, a student-teacher model for trimmed video classification is presented.
With reinforcement learning in~\cite{yeung2016end}, an agent predicts the next move.
Most recently, SCSampler~\cite{korbar2019scsampler} proposes a method for sampling salient segments in the untrimmed videos of Sports-1M~\cite{karpathy2014large}.
Conditioned on only the segment, it predicts a score for how salient this segment is to the action.

Conversely, in long-range activities, the importance of each segment is conditioned on not only the segment but also its context.
Thus, SCSampler is less suited for such activities.
This paper presents TimeGate, a novel selection method tailored for these activities.

\begin{figure*}[!ht]
\begin{center}
\includegraphics[trim=0mm 5mm 0mm 5mm,width=0.8\linewidth]{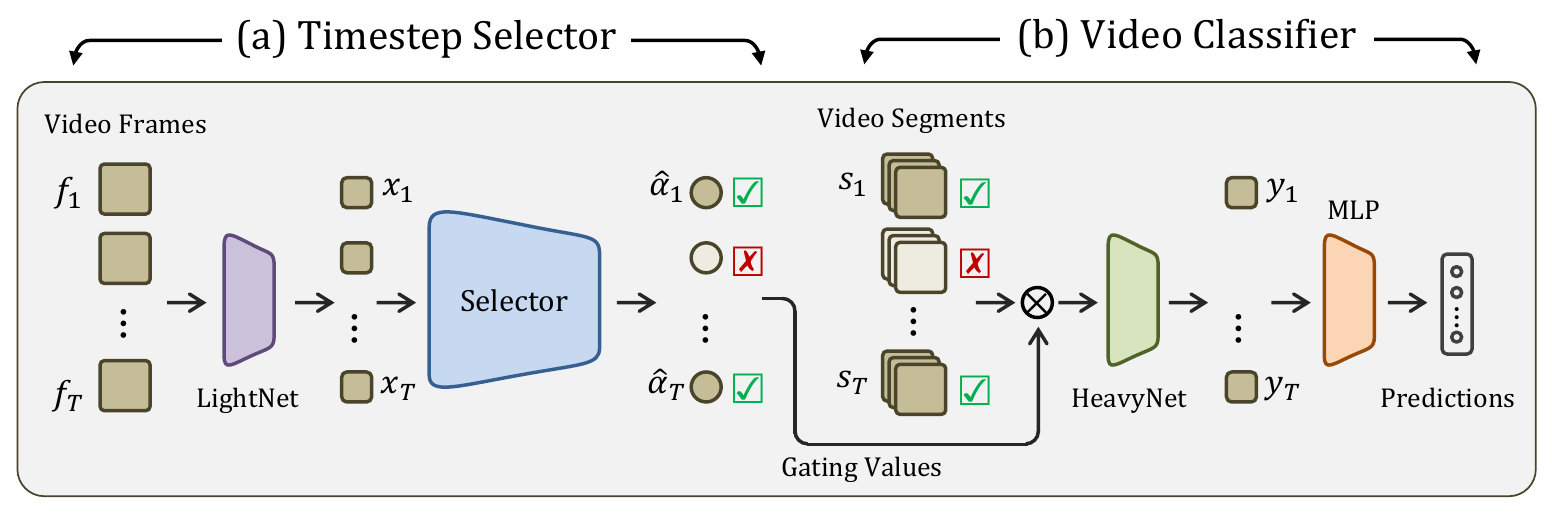}
\end{center}
\caption{Overview of the proposed model, TimeGate, with two stages.
The first stage is the timestep selector, left.
Based on a lightweight CNN, \textit{LightNet}, the model learns to select the most relevant timesteps for classifying the video.
This selection is conditioned on both the features of timestep and its context.
The second stage is the video classifier, right.
In which, only the selected timesteps (\textcolor{ForestGreen}{\CheckBoxCustom}) are considered, while the unselected timesteps (\textcolor{BrickRed}{\CrossedBox}) are completely ignored.
In this stage, a heavyweight CNN, \textit{HeavyNet} is used for feature representation of only the selected timesteps, followed by MLP for classification.}
\label{fig:3-1}
\end{figure*}

\section{Method}
\label{sec:method}
\subsection{TimeGate}

\partitle{Model Overview.}
TimeGate consists of two stages: timestep selector and video classifier, see figure~\ref{fig:3-1}.
The first stage is the selector, which consists of a lightweight CNN, \textit{LightNet}, followed by a novel gating module, see figure~\ref{fig:3-2}.
Its purpose is to select the most relevant timesteps from a minutes-long video.
The second stage is the classifier.
Its purpose is to learn deep and discriminatory video-level representations for maximum classification accuracy.
Thus, it resides on top of a heavyweight CNN, \textit{HeavyNet}, followed by a Multi-Layer Perceptron (MLP) for classification.
Only the timesteps chosen by the first stage, the timestep selector, are considered by the second stage, the video classifier.

\ptspace
\partitle{Timestep Selector.}
The selector takes as an input a uniformly sampled $T$ frames from a long-range video $v = \{ f_i \; | \; i \in [1, ..., T] \}$.
All the frames are represented as convolutional features  $ X = \{ x_i \; | \; i \in [1, ..., T] \}$, $ X \in \Ree^{ T \timesnarrow C \timesnarrow 1 \timesnarrow 1}$, where $C$ is the number of channels.
The objective of the selector is to choose only a few of these features.
In other words, we want to select only the timesteps that are most representative of the activity in the video, where each timestep is represented as a feature $x_i$.
Our hypothesis is that, a lightweight feature representation using an efficient CNN, \textit{LightNet}, would suffice for the selection.
Thus, the features $X$ are obtained from the last convolutional layer of the LightNet, and average-pooled globally over space, so the spatial dimensions of $X$ are $1 \timesnarrow 1$.

\ptspace
\partitle{Concept Kernels.}
The next step is to take binary decision of considering or discarding the timesteps.
But how to decide if a timestep feature $x_i$ is relevant or not?
Conceptually speaking, a long-range activity consists of few yet dominant and discriminative visual evidences, based on which, the video can be recognized~\cite{hussein2019timeception}.
Take for example ``making pancake''.
One can easily discriminate it by observing the evidences ``pancake'', ``eggs'', ``pan'', and ``stove''.
These evidences can be thought of as \textit{latent} concepts.
To represent them, we learn a set of concept kernels $ K =\{k_1, k_2, ...k_N\}, K \in \Ree ^ {N \times C}$, where  $N$ is the number of kernels, and $C$ is the kernel dimension.
$K$ are randomly initialized and are part of the network parameters.
They are learned during the training of the selector.
Our concept kernels $K$ are reminiscent of the centroids in ActionVlad~\cite{girdhar2017actionvlad}.

\ptspace
\partitle{Gating Module.}
The purpose of the gating module is to select the video timesteps, see figure~\ref{fig:3-2}, top.
The first step is to measure how relevant each timestep feature $x_i$ is to all of the concept kernels $K$ using an inner product $\odot$.
The result is the similarity vector $s_i = K^{\top} \odot x_i$, $s_i \in \Ree^{N \timesnarrow 1}$.
Our understanding is that the vector $s_i$ encodes how relevant a timestep is to each of $N$ concept kernels.
Then, based on this similarity vector $s_i$, we want to take a binary decision of considering or discarding the timestep feature $x_i$.
Therefore, we model the similarity vector $s_i$ using a two-layer MLP $f_\theta(\cdot)$.
The output layer of the MLP has a single neuron, denoted as $\alpha_i = f_{\theta}(s_i)$, $a_i \in \Ree^{1}$.

Intuitively, $\alpha_i$ is the gating decision corresponding to the timestep feature $x_i$.
Since $\alpha_i$ is a continuous variable, we cannot make a binary gating decision.
Thus, we make use of~\cite{jang2016categorical} to discretize $\alpha_i$ to binary gating variable $\hat{\alpha}_i$.
More formally, following the gating mechanism of~\cite{bejnordi2019batch}, we add gumbel noise $G$ to $\alpha_i$ and follow with \texttt{sigmoid} activation, thus $\hat{\alpha_i} = \texttt{sigmoid}(\alpha_i + G)$ .Then, we apply binary thresholding using the threshold value $\delta=0.5$.
So, we arrive at the binary gating value $\hat{\alpha}_i =\mathbb{I}_{(\delta > 0.5)} (\delta) $, $ \hat{\alpha}_i \in \{0, 1\}$, see figure~\ref{fig:3-2}, top.
Finally, for gating the $i$-th timestep, we multiply its feature $x_i$ with the binary value, resulting in the gated feature $ \hat{x_i} = x_i \cdot \hat{\alpha}_i$, $\hat{x_i} \in \Ree^{C \timesnarrow 1 \timesnarrow 1}$.

\begin{figure*}[t]
\minipage{0.24\textwidth}
\includegraphics[trim=3mm 8mm 3mm 5mm,width=\linewidth]{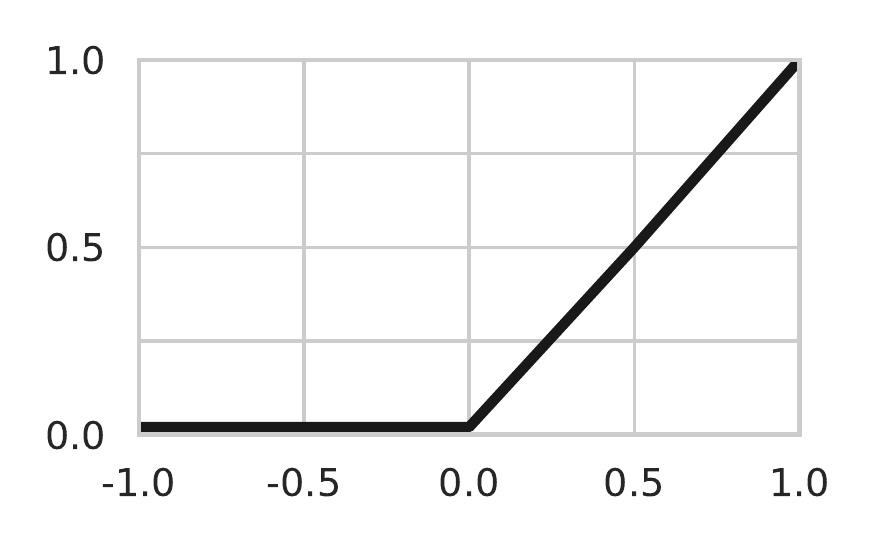}
\begin{center}
(a) ReLU
\end{center}
\endminipage\hfill
\minipage{0.24\textwidth}
\includegraphics[trim=3mm 8mm 3mm 5mm,width=\linewidth]{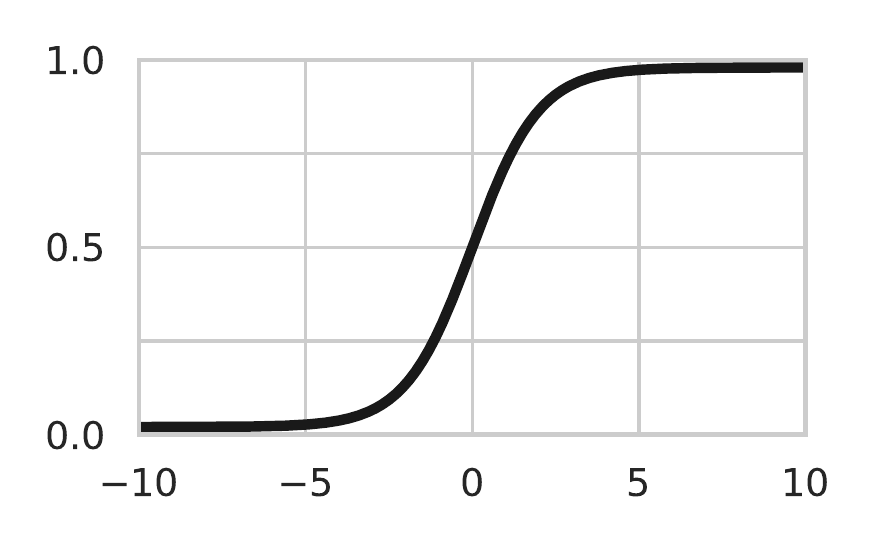}
\begin{center}
(b) Sigmoid
\end{center}
\endminipage\hfill
\minipage{0.24\textwidth}
\includegraphics[trim=3mm 8mm 3mm 5mm,width=\linewidth]{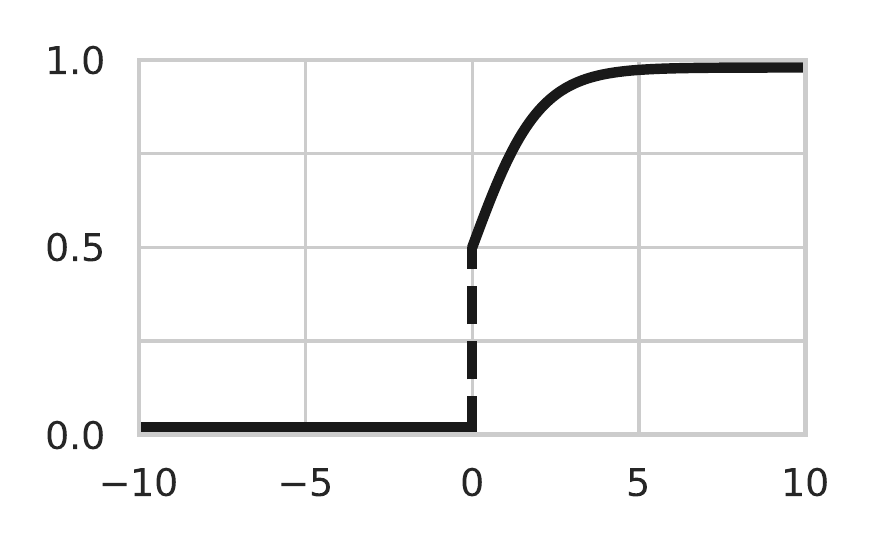}
\begin{center}
(c) Clipped sigmoid
\end{center}
\endminipage\hfill
\minipage{0.24\textwidth}
\includegraphics[trim=3mm 8mm 3mm 5mm,width=\linewidth]{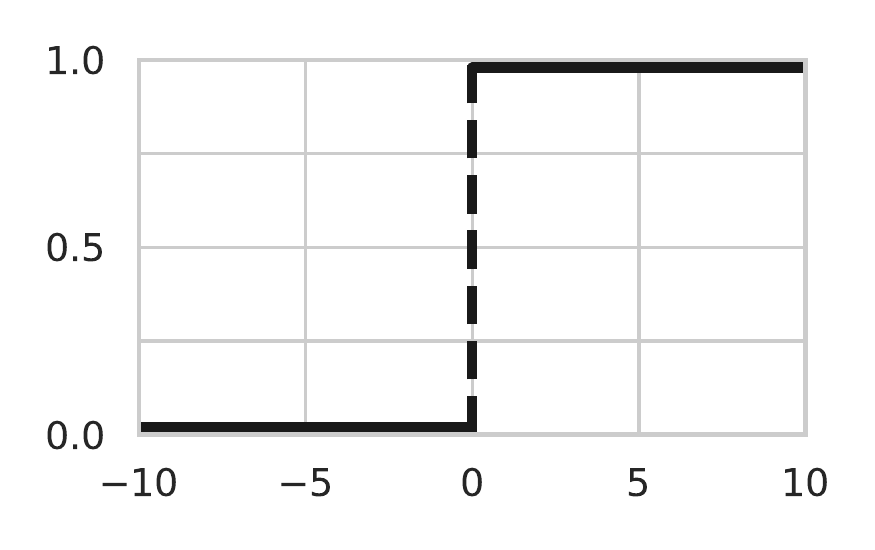}
\begin{center}
(d) Step Function
\end{center}
\endminipage
\vspace*{-2mm}
\label{fig:3-3}
\caption{
In training, we use \texttt{gated-sigmoid} to activate the gating value $\alpha_i$ and to select the timesteps.
\texttt{gated-sigmoid} has some desirable properties.
\textit{i}. Unlike \texttt{ReLU}, having upper bound does not allow a timestep feature to dominate others.
\textit{ii}. Different from \texttt{sigmoid}, being clipped allows the network to discard insignificant timesteps, \textit{i.e.} those with gating values $\alpha_i < 0.5$.
In test, we replace the \texttt{gated-sigmoid} with \texttt{step function} for binary gating of timesteps.}
\vspace*{-5mm}
\end{figure*}

\ptspace
\partitle{Gating Activation.}
A problem with using binary thresholding for gating, as in~\cite{bejnordi2019batch}, is that during training, the classifier does not know out of the gated timestep features, which is more relevant than the other.
Each $x_i$ is multiplied by a binary value $\hat{\alpha}_i \in \{0, 1\}$.
As a remedy, we propose \texttt{clipped-sigmoid} activation to replace the \texttt{sigmoid} activation used in~\cite{bejnordi2019batch}.
We find that this simply modified activation \texttt{clipped-sigmoid} is better suited for timestep gating due to three desirable properties, see figure 3.
\textit{i.} Being a relaxed version of the \texttt{step function} makes it differentiable.
\textit{ii.} Retaining the \texttt{sigmoid} value above the threshold means that the classifier gets the chance to know, out of the selected timesteps, which is relatively more important than the other.
\textit{iii.} Conversely to \texttt{ReLU}, the activation \texttt{clipped-sigmoid} is upper-bounded by one, thus preventing a single timestep feature $x_i$ from dominating the others by being multiplied by unbounded gating value $\hat{\alpha}_i$.

\begin{figure}[ht]
\begin{center}
\includegraphics[trim=0mm 5mm 0mm 2mm,width=0.7\linewidth]{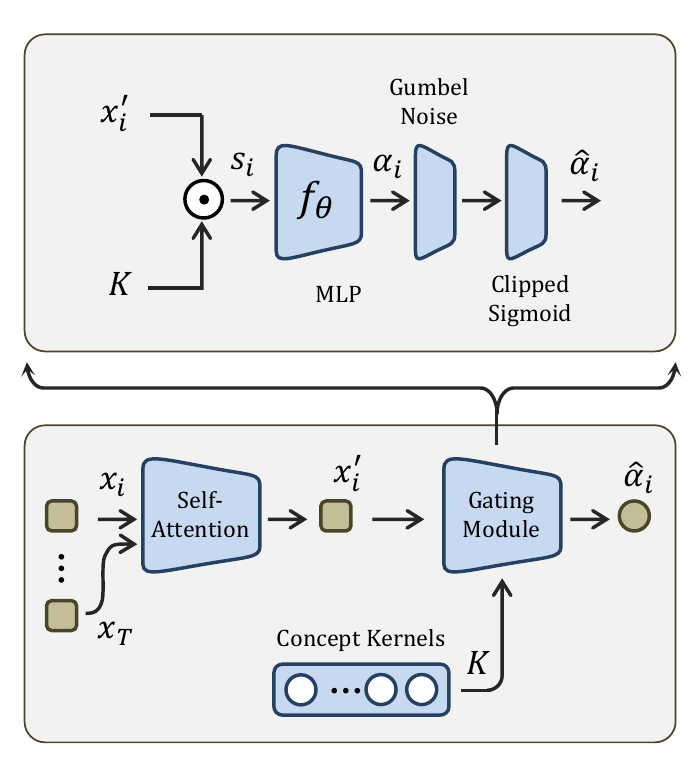}
\end{center}
\caption{Bottom, the timestep selector learns concept kernels $K$ to represent the most representative visual evidence.
Top, the gating module learns to select only a timestep feature $x_i$ according to its importance to the current video.}
\label{fig:3-2}
\end{figure}

\ptspace
\partitle{Context-Conditional Gating.}
Up till now, the selector learns to gate each timestep regardless of its context, \textit{i.e.} the other timesteps in the video.
To achieve context-conditional gating, where both the timestep and its context affect the gating decision, we opt for a temporal modeling layer, self-attention~\cite{wang2018non}, before the gating module, See figure~\ref{fig:3-2}, bottom.
This layer learns to correlate each timestep $x_i$ with all the others in the video $\{ x_1, ..., x_T\}$ before gating.

\ptspace
\partitle{Sparse Selection.}
The last component of the selector is to enforce sparsity on timestep selection, \textit{i.e.} choose as few timesteps as possible, yet retain the classification accuracy.
Simply put, the selector can cheat by predicting gating values just higher than the threshold $\alpha > \delta$, $\delta=0.5$, resulting in all gates opened and all timesteps selected.
The selector has a natural tendency to such a behaviour, as the only loss used so far is that of the classification.
And the more timesteps used by the classifier, the better the classification accuracy.
To prevent such a behaviour, we apply $L_0$ regularization~\cite{bejnordi2019batch,louizos2017learning} to all the gating values $\{ \hat{\alpha}_i \; | \; i \in [1, ..., T] \} $ to enforce sparsity on the selected timesteps.
We note that the sparsity regularization is necessary for a properly functioning gating mechanism.

\ptspace
\partitle{Video Classifier.}
The assumption of TimeGate is that having efficiently selected the most crucial timesteps from the video using the LightNet and the selector, one can opt for a much more powerful HeavyNet to effectively classify the video.
Thus, the second stage of TimeGate is the video classifier, see figure~\ref{fig:3-1}, left.
This classifier takes as input only the subset $T'$ of timesteps chosen by the selector, $T^{\prime} \subset T, \; T^{\prime} \ll T$.
Each timestep is represented as the feature of last convolutional layer of the HeavyNet.
The video-level features are denoted as $Y = \{ y_i \; | \, i \in [1, ..., T^{\prime}] \}$, $Y \in \Ree^{T^{\prime} \timesnarrow C^{\prime} \timesnarrow H \timesnarrow W}$, where $C^{\prime}$ is the number of channels, $T'$ is the number of selected timesteps, and $H, W$ are the spatial dimensions.
After the last convolutional layer, the video level features $Y$ are max-pooled over the spatial dimension and fed-forwarded to a two-layer MLP for classification. We follow~\cite{wang2018non} and max-pool the temporal dimension before the MLP \texttt{logits}.

\begin{figure*}[t]
\minipage{0.32\textwidth}
\includegraphics[trim=6mm 8mm 6mm 0mm,width=\linewidth]{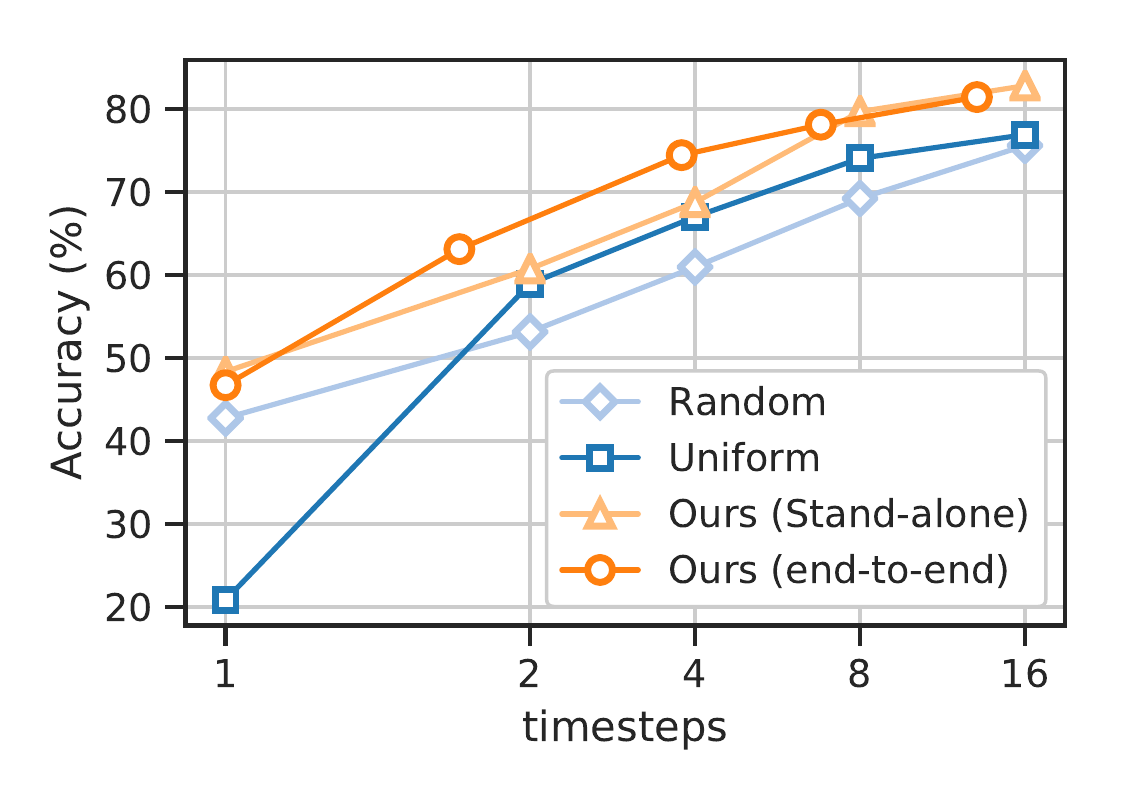}
\begin{center}
(a) I3D
\end{center}
\endminipage\hfill
\minipage{0.32\textwidth}
\includegraphics[trim=6mm 8mm 6mm 0mm,width=\linewidth]{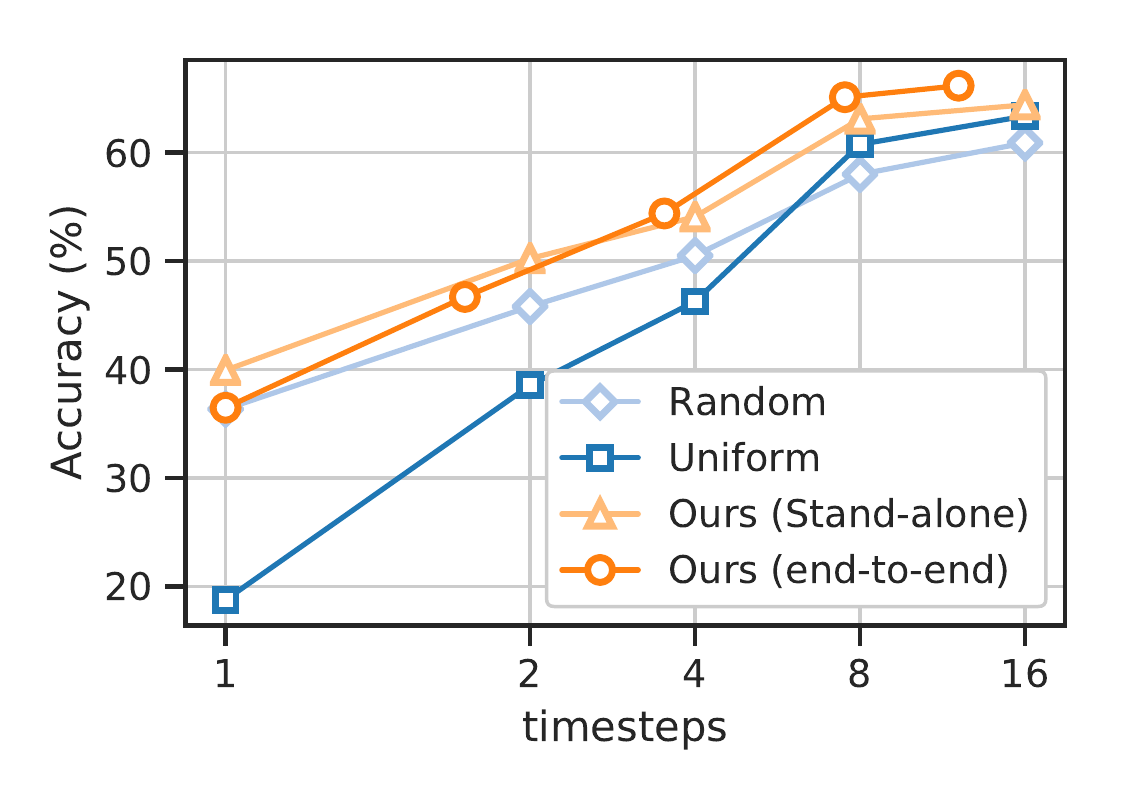}
\begin{center}
(b) ShuffleNet3D
\end{center}
\endminipage\hfill
\minipage{0.32\textwidth}
\includegraphics[trim=6mm 8mm 6mm 0mm,width=\linewidth]{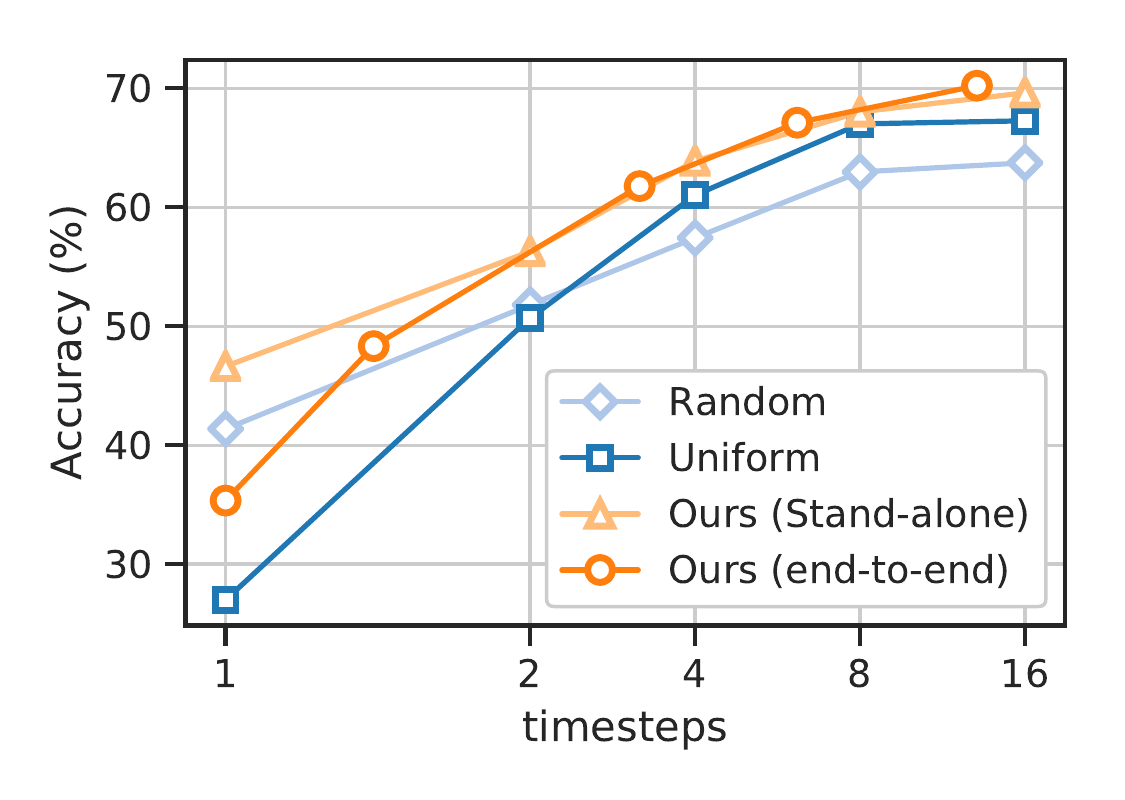}
\begin{center}
(c) ResNet2D
\end{center}
\endminipage
\vspace*{-2mm}
\caption{Our stand-alone timestep selector helps improving the performance and reduces the computation of off-the-shelf CNN classifiers -- be it 2D/3D heavyweight CNN or even lightweight 3D CNN.
More over, if TimeGate is trained end-to-end, the selector learns a better gating to the benefit of the classifier. So, the performance is improved even further.}
\label{fig:4-1}
\vspace*{-4mm}
\end{figure*}

\subsection{TimeGate Implementation}
\partitle{Backbone Choices.}
LightNet and HeavyNet are the backbone CNNs used by TimeGate.
Our choice for the LightNet is MobileNet-V3~\cite{sandler2018mobilenetv2}.
As for the HeavyNet, we explore three choices.
A powerful 3D CNN I3D~\cite{carreira2017quo}, an efficient 3D CNN ShuffleNet3D-V2~\cite{kopuklu2019resource}, and a powerful 2D CNN ResNet2D-50~\cite{he2016deep}.
Before training TimeGate on a specific dataset, the backbone CNNs are pre-trained on the dataset at hand.
We use the same training procedures specified by the authors of these CNNs.

\ptspace
\partitle{Timestep Alignment.} 
When the HeavyNet is a 3D CNN, the $i$-th timestep feature $y_i$ is obtained from processing the $i$-th video segment $s_i$ of $M$ successive frames $s_i = \{ f_{j}, ...., f_{j+M} \}$.
For I3D, $M=8$, and for ShuffleNet3D, $M=16$.
But since the LightNet of the selector is a 2D CNN, how can we align the timestep of the selector, with that of the classifier?
Simply put, for the aforementioned HeavyNet feature $y_i$, the aligned LightNet feature $x_i$ has to be obtained from the middle frame of the video segment $s_i$. More formally, the frame $f_{j + \lceil M/2 \rceil }$.

\ptspace
\partitle{Model Training.}
TimeGate is trained with batch size 32 and for 100 epochs.
We use Adam with learning rate $1e\text{-}3$ and epsilon $1e\text{-}4$.
We use PyTorch and TensorFlow for our implementation.
As for the number of concept kernels $N$, we found that $N=128$ is a good choice for all the experiments, similar to~\cite{hussein2019videograph}.
As for the gating module, see figure~\ref{fig:3-2}, during the training phase, we use gumbel noise and \texttt{clipped-sigmoid} to get the activated gating value $\hat{\alpha}_i$.
In the test phase, we don't use gumbel noise, and we replace \texttt{clipped-sigmoid} with \texttt{step-function}, to get the binary gating value $\hat{\alpha}_i =\mathbb{I}_{(\delta > 0.5)} (\delta) $.
That means alpha is binarized $ \hat{\alpha}_i \in \{0, 1\}$ with thresholding $\delta = 0.5$.

\section{Experiments}
\label{sec:experiments}

\subsection{Datasets}
\label{subsec:4-1}

\partitle{Charades}
is a widely used benchmark for human action recognition.
It is a diverse dataset with 157 action classes in total.
The task is mult-label recognition, where each video is assigned to one or more action class.
It is divided into 8k, 1.2k and 2k videos for training, validation and test splits, respectively, covering 67 hours.
On average, each video spans 30 seconds, and is labeled with 6 and 9 actions for training and test splits, respectively.
Thus, Charades meets the criteria of long-range activities.
We use Mean Average Precision (mAP) for evaluation.

\ptspace
\partitle{Breakfast}
is a benchmark for long-range activities, depicting cooking activities.
Overall, it contains 1712 videos, divided into 1357 and 335 for training and testing, respectively.
The task is video recognition into 10 classes of making different breakfasts.
Added to the video-level annotation, we are given temporal annotations of 48 unit-actions.
In our experiments, we only use the video-level annotation, and we do not use the temporal annotation of the unit-actions.
The videos are long-range, with the average length of 2.3 minutes per video.
Which makes it ideal for testing the efficiency of recognizing long-range activities.
The evaluation method is the classification accuracy.

\ptspace
\partitle{MultiThumos}
is a benchmark for long-range videos, depicting sports activities.
It consists of 413 videos, divided into 200 and 213 for training and testing, respectively.
Each video has multi-labels, where the total number of action classes across the dataset is 65.
The average length is 3.5 minutes per video.
The original task of this dataset~\cite{yeung2018every} is the temporal segmentation of these short-range actions.
Recently, it is repurposed by~\cite{hussein2019timeception} into multi-label classification of long-range videos.
We adopt the same experimental setup of~\cite{hussein2019timeception}.
That is to say, each long-range video is classified into multi-labels, and the mAP is used for evaluation.

\ptspace
\partitle{Ablation Studies.}
We use Breakfast as the primary dataset for the ablation experiments and studies.
These experiments highlight our contributions, as follows.
\textit{i.} In \textsection~\ref{subsec:4-3}, we discuss to what extend the end-to-end training of TimeGate helps.
\textit{ii}. In \textsection~\ref{subsec:4-4}, we show how context-conditional gating is more important than frame-conditional.
\textit{iii.} In \textsection~\ref{subsec:4-2},~\ref{subsec:4-5}, we demonstrate the improvements of TimeGate over the current CNN classifiers, in terms of accuracy and efficiency.

\subsection{Stand-alone Timestep Selector}
\label{subsec:4-2}

One might raise an important question
-- will a timestep selector based on LightNet features  $X$ benefit a classifier based on HeavyNet features $Y$, given the differences between the feature spaces of LightNet and HeavyNet $C \neq C^{\prime} $?
To answer this question, we construct an experiment of two steps on Breakfast.
The first step is training a stand-alone selector, where we choose MobileNet for both LightNet and HeavyNet.
During training, we randomly sample $T=32$ timesteps from each video.
Since MobileNet is a 2D CNN, a timestep here is practically a video frame.
With the help of the $L_0$ regularization, the selector achieves sparse selection of timesteps, by as little as $T^{\prime}=16$ without degrading the classification performance.
The second step is testing how will the selector benefit off-the-shelf CNN classifiers: I3D, ShuffleNet3D and ResNet2D.
Then, we measure their performance using different time scales.
More formally, from each test video,  we sample $T^{\prime}$ timesteps , $T^{\prime} \in \{1, 2, 4, 8, 16\}$, and we use different sampling methods: random, uniform, and timestep selector.
During testing, the output of the timestep selector is a per-timestep binary value $\hat{\alpha}_i \in \{0, 1\}$ of whether to consider or discard the $i$-th timestep.
So, if $T$ timesteps are processed by the selector, it is able to choose a subset $T'$ timesteps and discard the others, where $T' \subset T, T' \ll T$.
And to evaluate the benefit of the selector, the off-the-self classifier then uses only $T'$.

\begin{table}[!ht]
\centering
\renewcommand{\arraystretch}{1.0}
\setlength\tabcolsep{3.4pt}
\begin{tabular}{lccccccc}
\specialrule{0.3mm}{.0em}{.3em}
\multirow{2}{*}{Baseline} & \multicolumn{7}{c}{Accuracy (\%) @ Timesteps} \\
\cmidrule(lr){2-8}
 & 4 & 8 & 16 & 32 & 64 & 128 & 256 \\
\midrule
R2D                      & 61.0 & 67.1 & 67.3  & 71.0  & 72.9  &  74.3  &  73.8  \\
R2D+\textbf{TG}  & 63.9 & 68.2 & 70.2  & 73.3  & 74.3  &  \textbf{76.4}  & 74.3  \\
\midrule
S3D                      & 46.3 & 60.8 & 63.4 & 67.2 & 67.3 & 65.8 & 66.3  \\
S3D+\textbf{TG}  & 54.4 & 65.1 & 66.2 & \textbf{69.8} & 69.7 & 66.7 & 67.8 \\
\midrule
I3D                     & 66.8 & 74.3 & 82.8 & 84.7 & 85.7 &  86.5  &  85.4 \\
I3D+SCS~\cite{korbar2019scsampler} &  61.4  & 74.7 & 81.8 & 84.4 & 84.4 & 85.4 & 84.6 \\
I3D+\textbf{TG} & 69.5 & 77.9 & 85.2 & 85.9 & 86.7 & \textbf{88.1} & 86.5  \\
\specialrule{0.3mm}{.0em}{.0em}
\end{tabular}
\caption{The stand-alone selector of our model TimeGate (\textbf{TG}) benefits off-the-shelf CNN classifiers.
The benefit is consistent for various classifiers: I3D, ShuffleNet3D (S3D), and ResNet2D (R2D).}
\label{tbl:4-1}
\vspace*{-5pt}
\end{table}

As reported in table~\ref{tbl:4-1}, and shown in figure~\ref{fig:4-1}, we observe that the stand-alone selector improves the accuracy of off-the-shelf classifiers.
The reason is that the selector, based on LightNet, is able to select the most relevant timesteps from the video.
Also, we notice that the improvements are consistent for three different classifiers: I3D, ResNet2D and ShuffleNet3D.

\subsection{End-to-End TimeGate}
\label{subsec:4-3}
Having experimented with the stand-alone selector, we pose another question.
Is it possible to train TimeGate end-to-end, given that the selector and the classifier are based on two different CNNs, with two different feature spaces, $C \neq C^{\prime}$?.
Our experiments show that indeed, in end-to-end training, the gating module learns a better selection to the benefit of the classifier.
The outcome is improvement in performance over the stand-alone selection, as reported in table~\ref{tbl:4-4}.
We conclude that in end-to-end, the gating module learns to determine the importance of the $i$-th heavyweight feature $y_i$ based on the corresponding lightweight feature $x_i$.

\begin{table}[!ht]
\centering
\renewcommand{\arraystretch}{1.0}
\setlength\tabcolsep{3.0pt}
\begin{tabular}{lccccccc}
\specialrule{0.3mm}{.0em}{.3em}
\multirow{2}{*}{Baseline} & \multicolumn{7}{c}{Accuracy (\%) @ Timesteps} \\
\cmidrule(lr){2-8}
 & 4 & 8 & 16 & 32 & 64 & 128 & 256 \\
\midrule
SCSampler~\cite{korbar2019scsampler} &  61.4  & 74.7 & 81.8 & 84.4 & 84.4 & 85.4 & 84.6 \\
TimeGate \textbf{SA} & 69.5 & 77.9 & 85.2  & 85.9 &  86.7 & 88.1 & 86.5  \\
TimeGate \textbf{ETE} & 74.4 & 78.1 & 82.9 & 86.7 & 87.4 & \textbf{89.3} & 86.1 \\
\specialrule{0.3mm}{.0em}{.0em}
\end{tabular}
\caption{Our stand-alone (\textbf{SA}) selector benefits off-the-shelf CNN classifiers.
End-to-end (\textbf{ETE}) training is even better.}
\label{tbl:4-4}
\end{table}

\begin{figure}[!ht]
\begin{center}
\includegraphics[trim=5mm 18mm 5mm 2mm,width=1.0\linewidth]{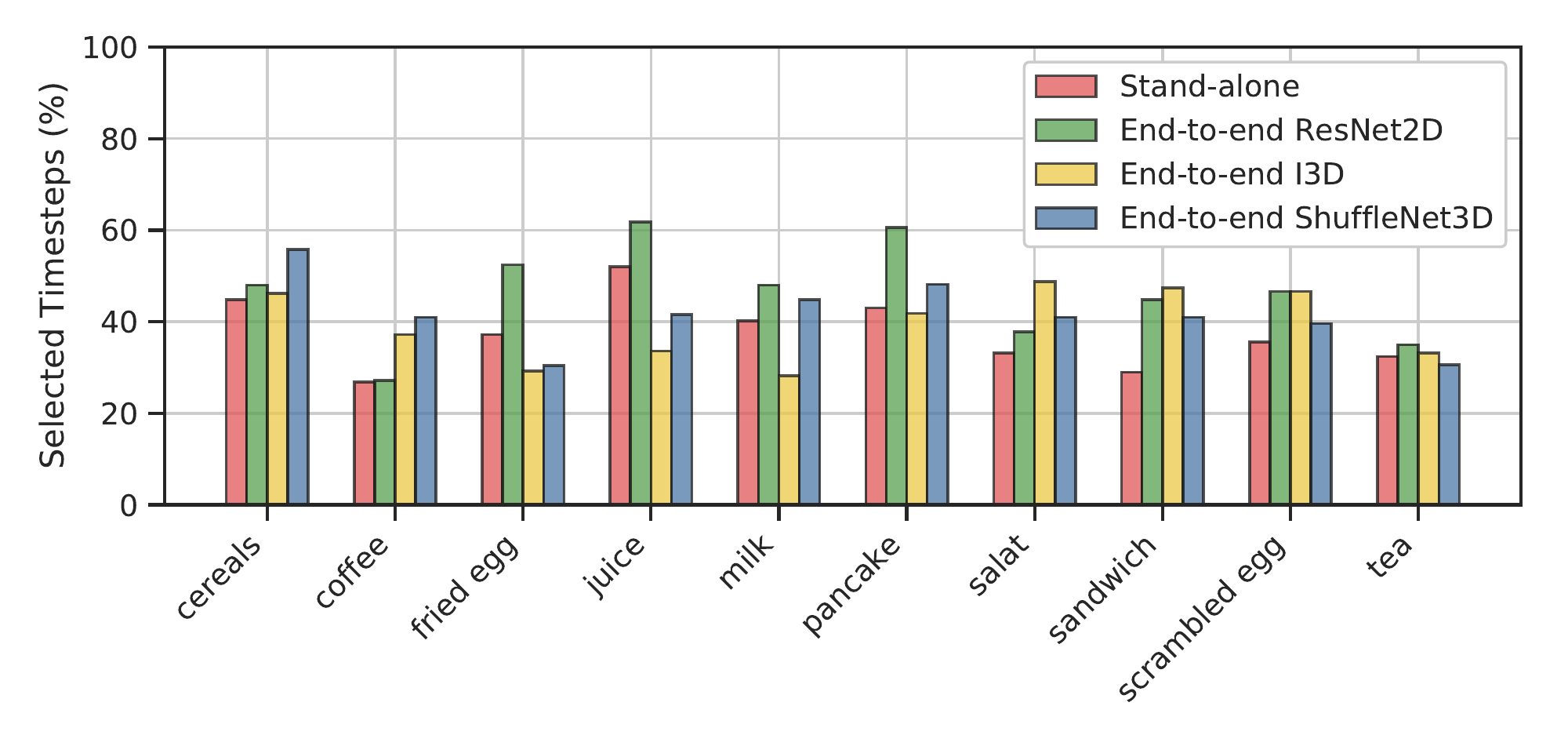}
\end{center}
\caption{
The ratios of selected timesteps for the activity classes of Breakfast.
Note the change in these ratios from stand-alone selector (red) to end-to-end training with the HeavyNets: ResNet2D (green) I3D (yellow) and ShuffleNet3D (blue).}
\label{fig:4-5}
\end{figure}

In addition, figure~\ref{fig:4-5} shows the average ratio of selected timesteps for each activity class of Breakfast dataset.
The ratios of the stand-alone (red) is changed when it is trained end-to-end with different HeavyNet: ResNet2D, (blue), I3D (yellow), and ShuffleNet3D (blue).
We observe that these ratios have similar trends when the HeavyNet is 3D CNN, regardless of which 3D CNN is used.
Between yellow and blue, there is a similar trend in 8 out of 10 activities.
However, these ratios tend to vary between 2D and 3D as HeavyNet -- only 3 out of 10 actions tend to have similar trends, see green and yellow.
From this experiment, we conclude that the gating module, depending on LightNet features, learns to select better timesteps to the benefit of the HeavyNet classifier.

\begin{table*}[!ht]
\centering
\renewcommand{\arraystretch}{1.0}
\setlength\tabcolsep{8.0pt}
\begin{tabular}{lcccccc}
\specialrule{0.3mm}{.0em}{.3em}
\multirow{2}{*}{} & \multicolumn{2}{c}{Timesteps} & \multicolumn{2}{c}{FLOPS (G) } & \multirow{2}{*}{Total FLOPS $\downarrow$}  & \multirow{2}{*}{Acc. (\%) $\uparrow$}  \\
\cmidrule(lr){2-3}
\cmidrule(lr){4-5}
 & LightNet & HeavyNet & LightNet+Gating & HeavyNet &  &  \\
\midrule
R2D                                         & ---  & 64    & --- & 246.6  & 246.6 & 72.9  \\
S3D+SCSampler~\cite{korbar2019scsampler}  & 128  & 16    & 7.5 & 61.7   & 69.2  & 68.6  \\
R2D+TimeGate                              & 128  & 16    & 7.8 & 61.7   & 69.5  & 70.2  \\
\midrule
S3D                                         & ---  & 64    & --- & 61.8   & 61.8  & 67.3  \\
S3D+SCSampler~\cite{korbar2019scsampler}  & 128  & 16    & 7.5 & 17.3   & 24.8  & 64.1  \\
S3D+TimeGate                              & 128  & 16    & 7.8 & 17.3   & 25.1  & 66.2  \\
\midrule
I3D                                         & ---  & 64   & ---  & 830.7  & 830.7 & 85.7  \\
I3D+SCSampler~\cite{korbar2019scsampler}  & 128  & 16   & 7.5  & 207.8  & 215.3 & 81.8  \\
I3D+TimeGate                              & 128  & 16   & 7.8  & 207.8  & 215.6 & 85.2  \\
\specialrule{0.3mm}{.0em}{.0em}
\end{tabular}
\caption{Breakdown of the computational cost of TimeGate \textit{v.s.} SCSampler.
Three choices of HeavyNet: ResNet2D (R2D), ShuffleNet3D (S3D) and I3D.
The computational cost of LightNet and the gating module is marginal compared to that of the HeavyNet.
TimeGate reduces the cost by almost half.
Our selector improves over SCSampler.}
\label{tbl:4-3}
\end{table*}

\subsection{Context-Conditional Gating}
\label{subsec:4-4}
When selecting the timesteps of long-range activities, TimeGate is conditioned on both the segment and its context.
This context-conditioning is an important novelty of TimeGate.
Also, this property is desired for long-range activities, because the importance of a certain segment is not always self-described, but rather depends on the context.

\begin{table}[!ht]
\centering
\renewcommand{\arraystretch}{1.0}
\setlength\tabcolsep{3.0pt}
\begin{tabular}{lccccccc}
\specialrule{0.3mm}{.0em}{.3em}
\multirow{2}{*}{Baseline} & \multicolumn{7}{c}{Accuracy (\%) @ Timesteps} \\
\cmidrule(lr){2-8}
 & 4 & 8 & 16 & 32 & 64 & 128 & 256 \\
\midrule
SCSampler~\cite{korbar2019scsampler} &  61.4  & 74.7 & 81.8 & 84.4 & 84.4 & 85.4 & 84.6 \\
TG Frame  & 69.2  &  73.8  & 80.7 & 81.5 & 83.9 & 83.1 & 83.6 \\
TG Context  & 69.5 & 77.9 & 85.2  & 85.9 &  86.7 & \textbf{88.1} & 86.5  \\
\specialrule{0.3mm}{.0em}{.0em}
\end{tabular}
\caption{TimeGate (TG) is better when the gating module is conditioned on both the frame-level and the context-level.
More over, TimeGate outperforms SCSampler in long-range activities.}
\label{tbl:4-2}
\end{table}

To validate this assumption, we design the following experiment.
We devise a baseline model of our timestep selector, that does not have a temporal layer before the gating module.
Thus, in this baseline, the gating is frame-conditioned.
Also, we include SCSampler~\cite{korbar2019scsampler} in this comparison.
We use I3D for the HeavyNet and we use MobileNet as the backbone CNN for both our timestep selector and SCSampler.
As reported in table~\ref{tbl:4-2}, we observe a drop in the performance when using the frame-conditioned TimeGate.
The reason is that, for long-range activities, its important for the selector to pay attention to the context of the video segment before sampling it.

\begin{figure}[!ht]
\begin{center}
\includegraphics[trim=3mm 13mm 5mm 5mm,width=1.0\linewidth]{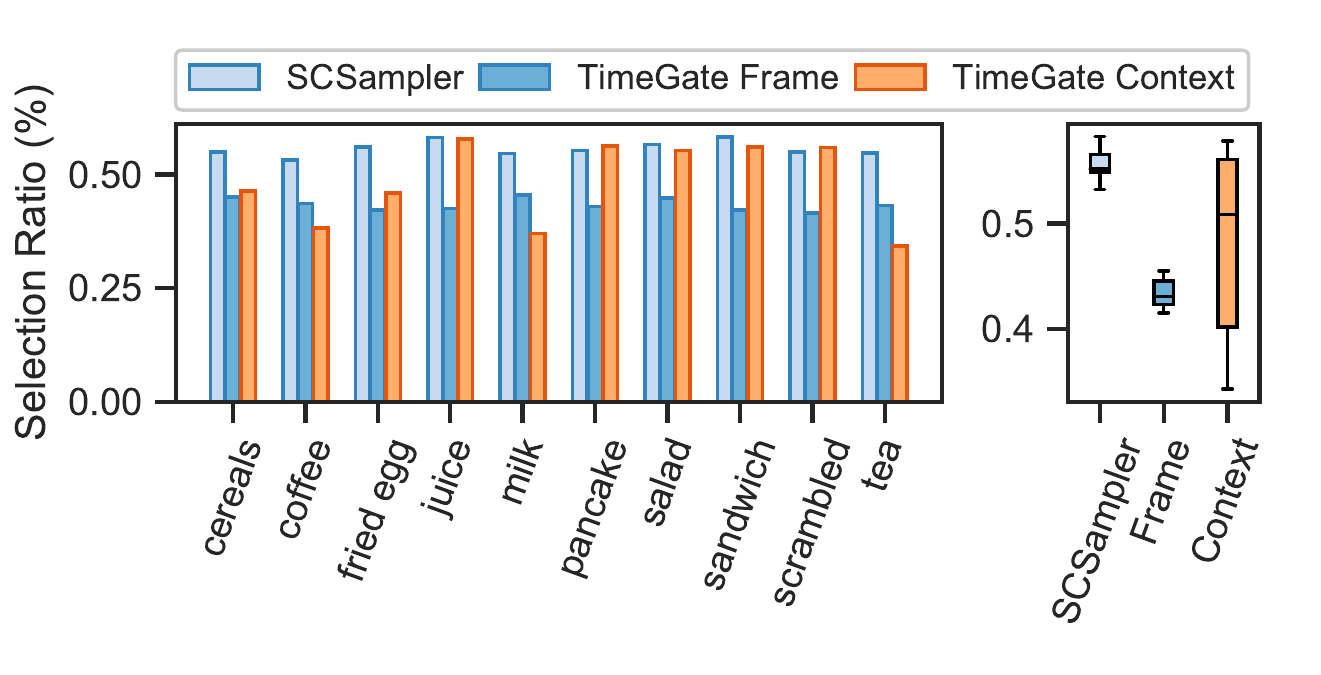}
\end{center}
\caption{In both the frame-conditioned TimeGate and SCSampler, the ratio of the selected timesteps have small variance across the activity classes of Breakfast.
In contrast, in context-conditioned TimeGate, the ratio is highly dependent on the activity, which means context-conditional gating is archived.}
\label{fig:4-6}
\end{figure}

We report another analysis in figure~\ref{fig:4-6}.
On the left, we show the ratio of selected timesteps for each activity class of Breakfast.
The frame-conditioned gating (dark blue) tends to select similar ratios regardless of the category, so does the SCSampler (light blue).
In contrast, we see more diverse ratios for the context-conditioned gating.
Figure~\ref{fig:4-6}, right, shows the ratio variances.
The much higher variance for context-conditional TimeGate means that it is more dependent on the activity class than the case of SCSampler or frame-conditional TimeGate.

\subsection{Computation-Performance Tradeoff}
\label{subsec:4-5}
\begin{figure}[!ht]
\begin{center}
\includegraphics[trim=10mm 12mm -10mm 5mm,width=0.8\linewidth]{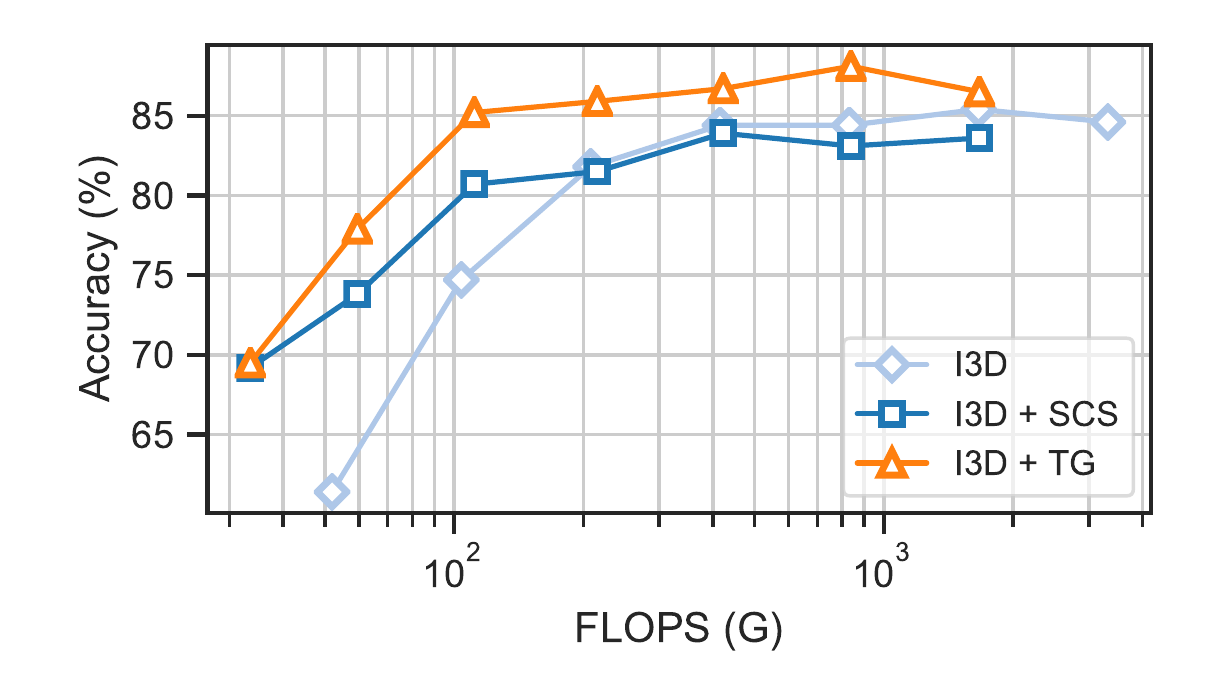}
\end{center}
\caption{TimeGate (\textbf{TG}) is better than SCSampler (\textbf{SCS}) in reducing computational cost of I3D.}
\label{fig:4-3}
\end{figure}

When it comes to the recognition of long-range activities, the golden rule is the more timesteps the better the accuracy, and the heavier the computation.
But given the huge redundancies of the visual evidences in these timesteps, there is a tradeoff between accuracy and computation.
In this experiment, we explore what is the effect of such a tradeoff on TimeGate, and we compare against SCSampler.
Figure~\ref{fig:4-3} shows this tradeoff using I3D as the video classifier.
We notice that both TimeGate and SCSampler can dramatically reduce the cost of I3D.
However, TimeGate outperforms SCSampler.

In table~\ref{tbl:4-3}, we report the exact computational budget of TimeGate \textit{v.s.} SCSampler and I3D.
We notice that with, carefully selected 16 timesteps out of 128, TimeGate is able to match the performance of off-the-shelf CNNs which use 64 uniformly sampled timesteps.
Also, we notice the computational cost of selecting these timesteps is marginal to the cost of the CNN classifier itself.
For example, to select 8 out of 128 Timesteps, TimeGate spends 7.5 G-FLOPS, while to classify only one timestep using I3D, 3.9 G-FLOPS are needed.

\subsection{Experiments on Charades}
\label{subsec:4-6}

\begin{figure}[!ht]
\begin{center}
\includegraphics[trim=5mm 10mm -5mm 5mm,width=0.9\linewidth]{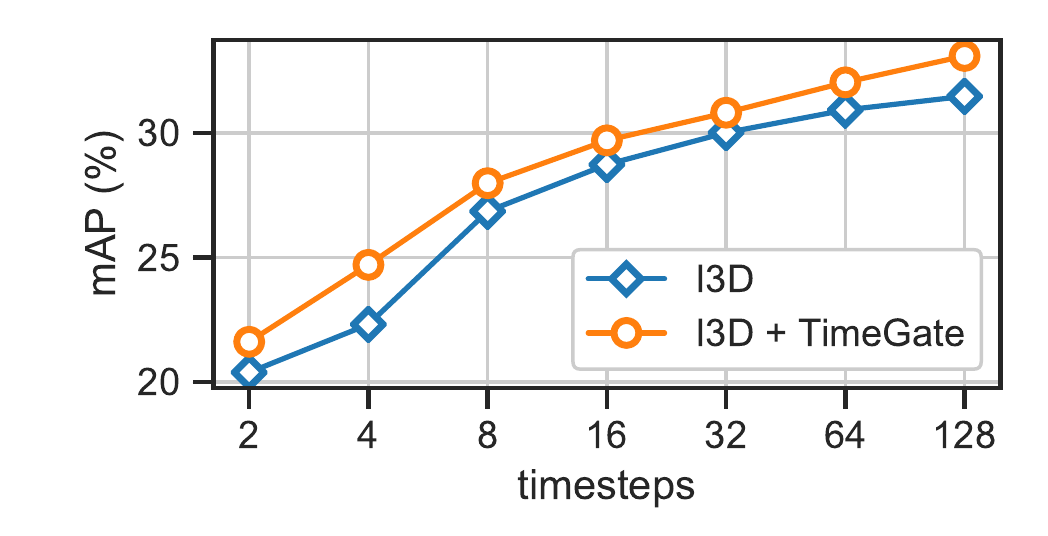}
\end{center}
\caption{TimeGate improves the performance of the off-the-shelf I3D for recognizing the actions of Charades.}
\label{fig:4-4}
\end{figure}

In this experiment, we test how TimeGate would fair against off-the-shelf CNN for recognizing the multi-label action videos of Charades.
This dataset is different from Breakfast in two ways.
First, the videos are mid-range with average length of 0.5 minutes, compared to 2 minutes of Breakfast.
Second, it is multi-label classification, but breakfast is single-label classification.
So, it is more challenging to select unrelated timesteps from the videos of Charades than Breakfast.
Most of the timesteps are already relevant to recognizing the mid-range videos of Charades.
Still, TimeGate outperforms I3D at different time scales, see figure~\ref{fig:4-4} and table~\ref{tbl:4-5}.

\begin{table}[!ht]
\centering
\renewcommand{\arraystretch}{1.0}
\setlength\tabcolsep{2.8pt}
\begin{tabular}{lccccccc}
\specialrule{0.3mm}{.0em}{.3em}
\multirow{2}{*}{Baseline} & \multicolumn{7}{c}{mAP (\%) @ Timesteps} \\
\cmidrule(lr){2-8}
 & 4 & 8 & 16 & 32 & 64 & 128 & 256 \\
\midrule
I3D            & 20.4 & 22.3 & 26.8 & 28.3 & 30.1 & 30.9 & 31.5 \\
I3D + TimeGate & 21.6 & 24.7 & 27.9 & 29.7 & 30.8 & 32.4 &  \textbf{33.1} \\
\specialrule{0.3mm}{.0em}{.0em}
\end{tabular}
\caption{TimeGate improves the performance of the backbone CNNs (\textit{i.e.} I3D) on the challenging task of multi-label classification of Charades.}
\label{tbl:4-5}
\end{table}

Worth mentioning that TimeGate consistently improves the efficiency of HeavyNet CNNs other than I3D.
For example, if TimeGate uses 3D-ResNet-101~\cite{wang2018non} as the HeavyNet, we achieve 36.2\% using 256 timesteps compared to 35.5\% achieved by~\cite{wang2018non} using dense sampling of 1024 timesteps.
In other words, TimeGate retains the performance of 3D-ResNet-101 using only 25\% of the computation.
The reason is that, when TimeGate selects the most relevant segments from each video, it improves the signal-to-noise ratio.
In analogy, ~\cite{korbar2019scsampler} concluded that when the CNN video classifier considers the unrelated video segments, the accuracy degrades.

\begin{figure*}[!ht]
\begin{center}
\includegraphics[trim=0mm 2mm 0mm 5mm,width=1.0\linewidth]{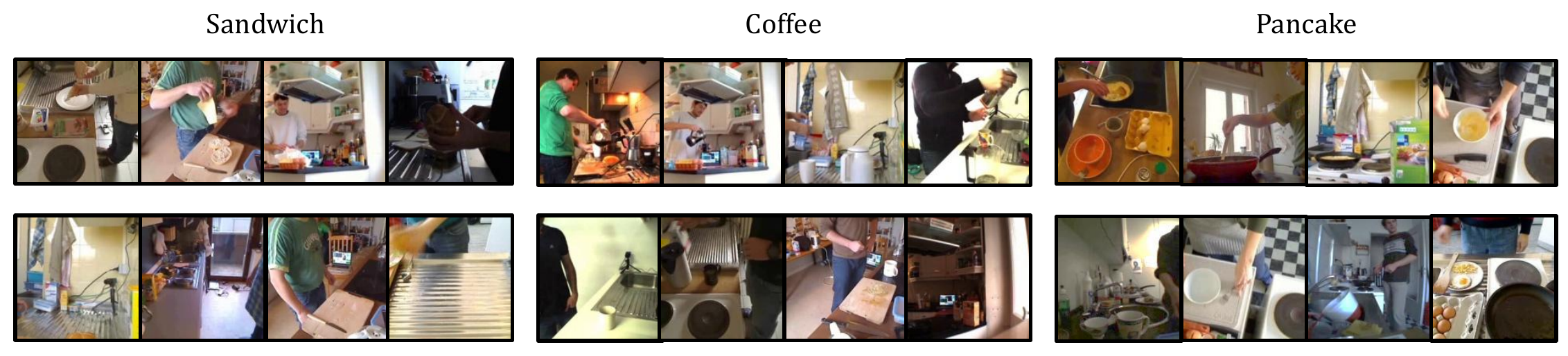}
\end{center}
\caption{Top, frames corresponding to the selected timesteps by TimeGate. Bottom, are those discarded by TimeGate.
The shown figures are for three activities: ``making sandwich", ``preparing coffee", and ``making pancake".
The general observation is that TimeGate tends to discard the segments with little discriminative visual evidences.}
\label{fig:4-9}
\end{figure*}

\subsection{Experiments on MultiThumos}
\label{subsec:4-7}
Our final experiment is to use TimeGate in classifying the long-range activities of MultiThumos.
This dataset is particularly challenging, as each video is multi-labeled.
Nevertheless, we observe that TimeGate is able to retain the performance of the HeavyNet (I3D) with much reduced computation, see table~\ref{tbl:4-9}.
In addition, it outperforms SCSampler in reducing the computational cost.
Worth mentioning that TimeGate achieves 75.11\% mAP using 256 timesteps compared to 74.79\% mAP achieved by~\cite{hussein2019timeception} using dense-sampling of 1024 timesteps.
In other words, TimeGate retains the performance of~\cite{hussein2019timeception} with almost 25\% of the computational cost.

\begin{table}[!ht]
\centering
\renewcommand{\arraystretch}{1.0}
\setlength\tabcolsep{2.8pt}
\scalebox{0.85}{
\begin{tabular}{lccccccc}
\specialrule{0.3mm}{.0em}{.3em}
\multirow{2}{*}{Baseline} & \multicolumn{7}{c}{mAP (\%) @ Timesteps} \\
\cmidrule(lr){2-8}
 & 4 & 8 & 16 & 32 & 64 & 128 & 256 \\
\midrule
I3D             & 41.85 & 45.02 & 52.75 & 58.41 & 64.74 & 67.19 & 69.32 \\
I3D + SCSampler & 43.51 & 47.68 & 54.14 & 60.87 & 67.23 & 69.83 & 72.46 \\
I3D + TimeGate  & 45.38 & 50.02 & 57.63 & 63.34 & 69.07 & 73.20 & \textbf{75.11} \\
\specialrule{0.3mm}{.0em}{.0em}
\end{tabular}}
\vspace*{5pt}
\caption{TimeGate improves the performance of I3D when classifying the long-range activities of MultiThumos.
Also, it outperforms SCSampler.}
\label{tbl:4-9}
\end{table}

\subsection{Qualitative Results}
\label{subsec:4-8}
\partitle{Examples of Gated Timesteps.}
In figure~\ref{fig:4-9}, we show a few visual examples of the timesteps selected, top, and discarded, bottom, by the gating module.
We consider three activities: ``making sandwich", ``preparing coffee", and ``making pancake".
The general observation is that TimeGate tends to discard the segments with little discriminative visual evidences.

\begin{figure}[!ht]
\begin{center}
\includegraphics[trim=2mm 14mm 2mm 4mm,width=1.0\linewidth]{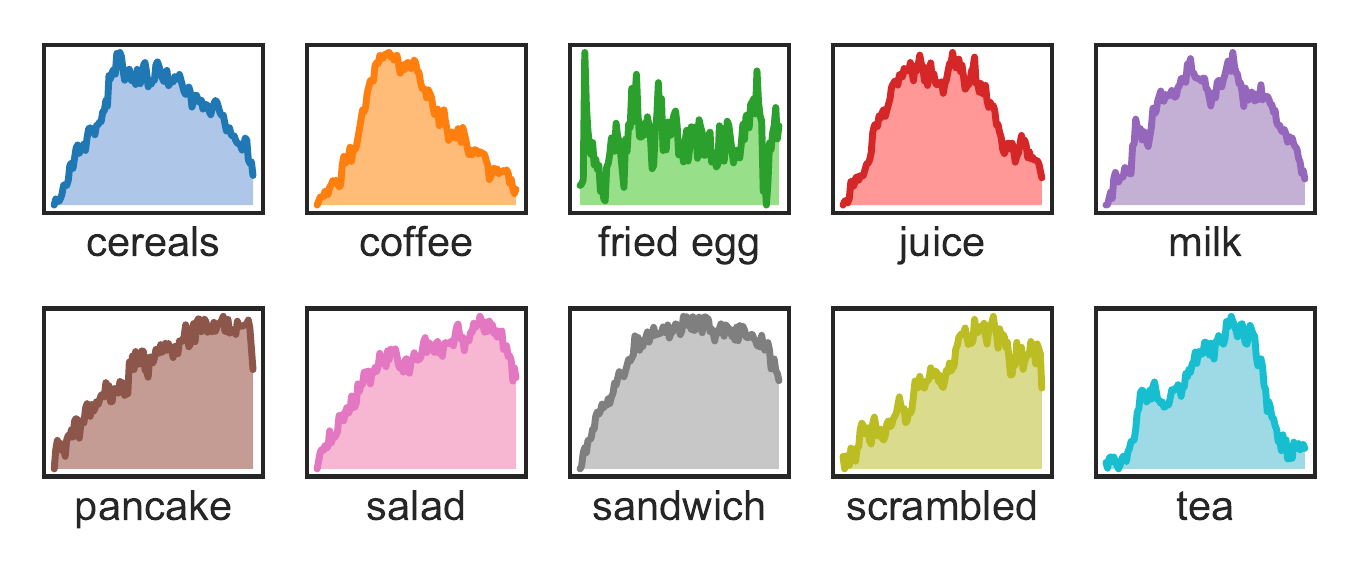}
\end{center}
\caption{Distribution of the gating values across time for each activity of Breakfast.
In simple activities, such as ``making coffee", most of the selected segments happen in the middle of the video.
This means the middle of the video is much more relevant than the other parts.
While in complex activities, such as ``making sandwich", the selected segments tend to distribute across the entire video.
This means that almost the entire video contains relevant and important segments.}
\label{fig:4-7}
\end{figure}

\ptspace
\partitle{Distribution of Gating Values.}
One might ask the question, how evenly distributed are the timesteps selected by TimeGate?
To answer this question, we uniformly sample $T=128$ timesteps from each test video.
Then, we predict the gating value $\alpha_i$ for each timestep.
After that, for all the videos of the same activity class, we average their gating values.
Next, we normalize these values between zero and one, and visualize them in figure~\ref{fig:4-7}.
Our observation is that, some activities are simple and usually happen in the middle of the video, such as ``preparing tea", or ``making coffee".
Others are complex and occupy the entire video, such as ``fried egg" or ``making sandwich".

\section{Conclusion}
\label{sec:conclusions}
In this paper, we proposed TimeGate, a neural model for efficient recognition of long-range activities in videos.
Our approach for realizing the efficiency is sampling the most relevant segments from the activity video.
We highlighted the differences between sampling for short-range actions \textit{v.s.} long-range activities.
We also stated the limitations of existing works, such as SCSampler.
TimeGate overcomes these limitations using three contributions.
First, a differentiable gating module for timestep selection.
Second, the selection that is conditioned on both the timestep and its context.
Third, TimeGate, an end-to-end neural model to retain the performance of existing CNN classifiers at a fraction of the computational budget.
We experimented on three benchmarks and compared against related works. 
TimeGate consistently outperforms competing methods on all three benchmarks and reduces the computation of  I3D  by 50\% while maintaining the classification accuracy. On MultiThumos, TimeGate sets a new state-of-the-art mAP of 75.11\% compared to 74.79\% mAP of Timeception~\cite{hussein2019timeception} while consuming only 25\% of the computation cost.
Our empirical evaluations and results demonstrate the efficiency of TimeGate in recognizing long-range activities.

{\small
\bibliographystyle{unsrt}
\bibliography{main}}

\end{document}